\documentclass[APA,STIX1COL]{WileyNJD-v2}
\usepackage[section]{placeins}

\articletype{}%

\begin{document}

\title{SERT: A Transfomer Based Model for Spatio-Temporal Sensor Data  with Missing Values for Environmental Monitoring}

\author[1]{Amin Shoari Nejad*}

\author[2]{Rocío Alaiz-Rodríguez}

\author[3]{Gerard D. McCarthy}

\author[4]{Brian Kelleher}

\author[4]{Anthony Grey}

\author[1]{Andrew Parnell}

\authormark{SHOARI NEJAD \textsc{et al}}

\address[1]{Hamilton Institute, Insight Centre for Data Analytics, Maynooth University, Kildare, Ireland}

\address[2]{Department of Electrical, Systems and Automation, University of León, Spain}

\address[3]{
ICARUS, Department of Geography, Maynooth University, Maynooth, Ireland}

\address[4]{
Organic Geochemical Research Laboratory, Dublin City University, Glasnevin Campus, Dublin 9, Ireland}

\corres{*Amin Shoari Nejad, \email{amin.shoarinejad@gmail.com}}

\abstract[Abstract]{Environmental monitoring is crucial to our understanding of climate change, biodiversity loss and pollution. The availability of large-scale spatio-temporal data from sources such as sensors and satellites allows us to develop sophisticated models for forecasting and understanding key drivers. However, the data collected from sensors often contain missing values due to faulty equipment or maintenance issues. The missing values rarely occur simultaneously leading to data that are multivariate misaligned sparse time series. We propose two models that are capable of performing multivariate spatio-temporal forecasting while handling missing data naturally without the need for imputation. The first model is a transformer-based model, which we name \textbf{SERT} (\textbf{S}patio-temporal \textbf{E}ncoder \textbf{R}epresentations from \textbf{T}ransformers). The second is a simpler model named \textbf{SST-ANN} (\textbf{S}parse \textbf{S}patio-\textbf{T}emporal \textbf{A}rtificial \textbf{N}eural \textbf{N}etwork) which is capable of providing interpretable results. We conduct extensive experiments on two different datasets for multivariate spatio-temporal forecasting and show that our models have competitive or superior performance to those at the state-of-the-art.}

\keywords{Spatio-temporal, Deep learning, Transformers, Environmental monitoring}

\maketitle

%--------------------------------Introduction----------------------------------------

\section{Introduction}\label{intro}
The importance of spatio-temporal forecasting has increased significantly in recent years due to the availability of large-scale spatio-temporal data from various sources such as sensors and satellites \citep{hamdi2022spatiotemporal}. Spatio-temporal forecasting involves predicting how data vary over space and time, which is critical for a wide range of applications such as water quality forecasting \citep{deng2022spatiotemporal}. A common approach to modelling spatio-temporal data is to use a multivariate time series structure, where each time series is associated with a variable at a specific location \citep{wikle2015}.

\begin{figure}[htbp]
\centerline{\includegraphics[width=0.7\textwidth]{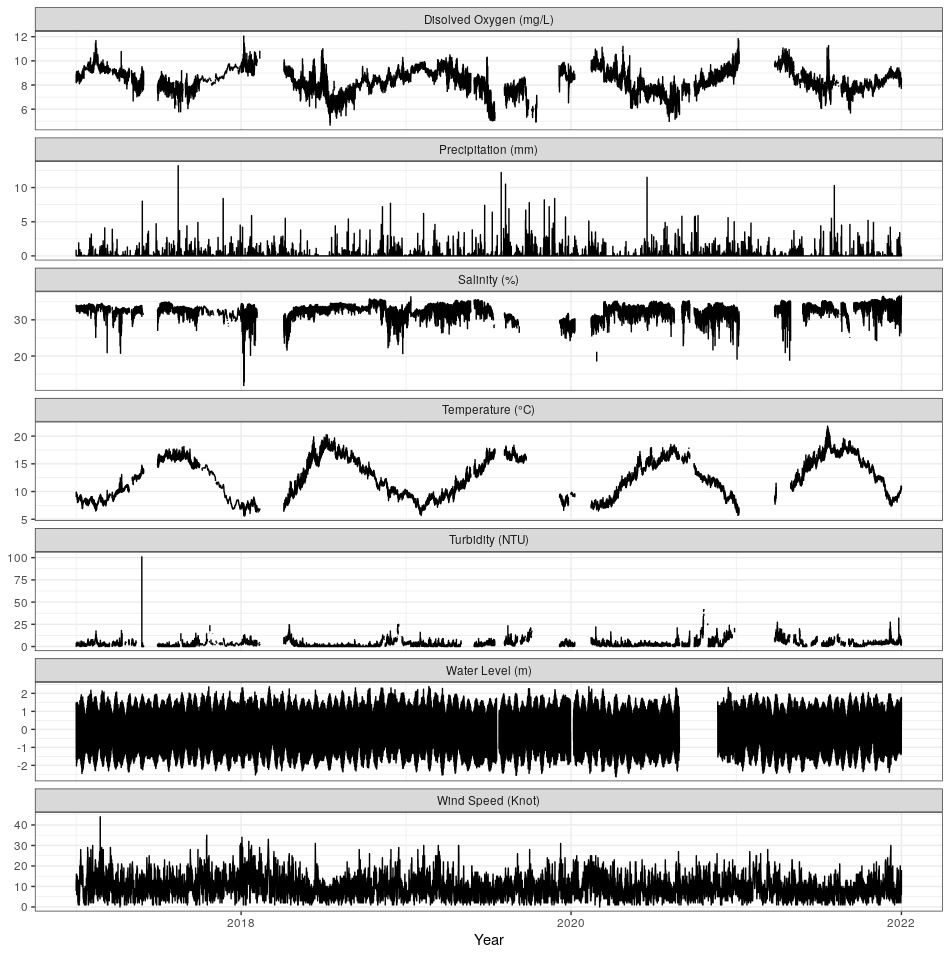}}
\caption{An example of our environmental monitoring dataset; a multivariate environmental time-series with missing values.}
\label{fig:rawdata}
\end{figure}

Spatio-temporal data often contain missing values which is a common problem in environmental monitoring, and can be caused by sensor failure, malfunction or communication problems (for example see Figure \ref{fig:rawdata}). A common remedy for forecasting with missing data is to impute the missing values using a variety of methods \citep{van2018flexible, stekhoven2012missforest}. However, these methods are not always effective, can reduce the signal in the data, and thus make the modelling or forecasting task more challenging and unreliable. Therefore it is important to design models that can handle missing data naturally whilst simultaneously learning the underlying patterns in the data without resorting to imputation.

We propose a new model that is capable of performing multivariate spatio-temporal forecasting named SERT (\textbf{S}patio-temporal \textbf{E}ncoder \textbf{R}epresentations from \textbf{T}ransformers). Our model is an extension of the well-known transformer architecture that has shown remarkable success in natural language processing and also image analysis \citep{devlin2018, dosovitskiy2020image}. SERT is designed to capture the complex joint temporal and spatial dependencies among the input variables. An important feature of our proposed model which differentiates it from the many other available methods is its ability to handle missing data more naturally without requiring any missing value imputation. 

In addition to the SERT model, we introduce an interpretable simplified version that provides insights into the underlying factors that drive the predicted values, which can assist in decision and policy-making. Our proposed simplified model is named SST-ANN (\textbf{S}parse \textbf{S}patio-\textbf{T}emporal \textbf{A}rtificial \textbf{N}eural \textbf{N}etwork) and removes the transformer layers from the SERT structure. Despite being less accurate than SERT, it is capable of providing insightful results with faster computation time while similarly being able to handle missing values. Depending on the complexity of the problem, the required accuracy and the available computational resources, the user can fit both SERT and SST-ANN, using the former to provide more accurate forecasts and the latter to forecast and gain insights about how the results were obtained. 

To evaluate the performance of our proposed models, we conducted extensive experiments on two different datasets for multivariate spatio-temporal forecasting. We fitted our models to a simulated dataset to assess their ability to function under different levels of sparsity. We also evaluated the performance of the models on a real-world dataset, including missing values, of environmental variables in Dublin Bay for 7 hour ahead forecasting. Our experimental results show that our models are competitive with state-of-the-art models for multivariate spatio-temporal forecasting.

Our paper is organized as follows. In Section \ref{litrev}, we provide a brief overview of related work on models developed for analysing sequential data in general and spatio-temporal forecasting applied to environmental monitoring in particular. In Section \ref{model}, we describe the proposed SERT and SST-ANN models in detail. In Section \ref{experiments}, we present the experimental results and analysis. Finally, in Section \ref{conclusion}, we conclude the paper and discuss future directions of research.

%--------------------------------Related Works-------------------------------------

\section{Related Work}\label{litrev}
In this section we provide a brief overview of the recent developments in deep learning models for sequential data analysis and spatio-temporal models for environmental monitoring, and also methods for handling missing data and adding interpretability to deep learning models applied to time series data.

\subsection{Deep Learning Models for Sequential Data}\label{dl_seq}
Recurrent Neural Networks (RNNs) have been one of the most popular deep learning models for sequential data \citep{li2018independently}. However, RNNs suffer from the vanishing gradient problem which makes them unable to learn long-term dependencies in the data. To address this problem, Long Short-Term Memory (LSTM) networks \citep{hochreiter1997} and Gated Recurrent Unit (GRU) networks \citep{cho2014} were introduced. These models have been applied to various tasks such as machine translation \citep{bahdanau2014}, speech recognition \citep{graves2013}, and time series forecasting \citep{lim2021time}.

More recently a new type of deep learning model named transformers was introduced \citep{vaswani2017}. Transformers are based on the attention mechanism which enables them to learn the dependencies between the input and output sequences. Transformers are comprised of an encoder and a decoder network. The encoder network is responsible for learning the representation of the input sequence, while the decoder network is responsible for generating the output sequence based on the learned representation. Models developed on the transformer architecture include BERT \citep{devlin2018}, which uses only the encoder part, and GPT \citep{radford2018} which uses only the decoder part. These models have been applied to various tasks in natural language processing such as question answering \citep{devlin2018}, text classification \citep{sun2019}, and text summarization \citep{liu2019}. Overall, transformers have proven themselves to be more effective than recurrent based models in many applications, especially in natural language processing.

\subsection{Deep Learning Models for Spatio-Temporal data}\label{dl_st}

Spatio-temporal forecasting is often framed as multivariate time series forecasting where each time series is associated with a variable at a specific location. The application of deep learning models for spatio-temporal forecasting is not new. For example, \cite{Zhang2018} used an LSTM to forecast daily land surface temperature, and \cite{mcdermott2017} developed an ensemble quadratic echo state network for forecasting Pacific sea surface temperature. However, the application of transformers to spatio-temporal forecasting is relatively new and challenging because transformers have been mainly developed in the field of natural language processing (NLP). Nonetheless researchers were inspired by the success of transformers in NLP and started adapting them to spatio-temporal forecasting which can be formulated as a sequence-to-sequence problem, where the input is a sequence of historical observations of multiple variables at different locations, and the output is a sequence of future predictions of the same variables at the same locations. A common approach for sequence-to-sequence modeling is to use an encoder-decoder architecture, where an encoder network maps the input sequence into a latent representation, and a decoder network generates the output sequence from the latent representation \citep{sutskever2014sequence}. \cite{grigsby2019spatiotemporal} used this idea to develop a new model called Spacetimeformer and applied it to traffic prediction and weather forecasting. However, to the best of our knowledge, the application of transformers to environmental monitoring is limited to the recent work by \cite{yu2023} who used a transformer-based model for hourly $PM_{2.5}$ forecasting in Los Angeles.

\subsection{Addressing Missing Values in Modelling}\label{missing}

As mentioned in the introduction, a major challenge in spatio-temporal forecasting in the environmental monitoring context is dealing with missing values. A common approach for dealing with the missing values is imputation \citep{van2018flexible} before conducting any analysis. In time series modelling care needs to be taken to avoid introducing bias; last observation carried forward is a common approach. An alternative used in the literature is that of a Bayesian framework which enables defining a prior distribution over the missing values so that they can be inferred with the other unobserved parameters when fitting the models. For example, \citep{shoarinejad2022} proposed a Bayesian model (called VARICH) for spatio-temporal modelling of turbidity data with many missing values. However, this approach is computationally expensive and requires a large number of samples from the posterior distribution to obtain acceptable results and thus is not suitable for large spatio-temporal datasets.

To address the missing data problem in time series, \cite{horn2020SeFT} proposed a novel approach to encode multivariate time series using set functions and introduced a new model called SeFT for classifying time series with irregularly sampled clinical data. More recently, \cite{Tipirneni2022STraTS} proposed a new transformer based model called STraTS that represents each observation as a triplet of the form (time, variable name, value). As opposed to SeFT, STraTS uses a learnable positional encoding and a Continuous Value Embedding (CVE) scheme that is a one-to-many feed-forward network. STraTS was developed to perform multivariate time series forecasting during the pre-training phase and classification as the final task on irregularly sampled clinical data. It has been shown to have higher accuracy than its predecessor, the SeFT model, when applied to the classification of clinical time series. Both SeFT and STraTS can handle missing values without requiring any imputation. To the best of our knowledge, none of these novel methods have been applied to environmental monitoring challenges; we adapt STraTS to a spatio-temporal setting. 

\subsection{Interpretability of Deep Learning Models}\label{interpretability}

Deep learning models are often considered as black-box models because they are hard to interpret \citep{buhrmester2021analysis}. However, in many applications, it is important to understand the model's decision making process \citep{du2019techniques}. Some authors have proposed methods to help interpret the results of deep learning models applied to time series data. For example, SeFT uses the attention mechanism in its architecture and the authors of the work showed that the attention weights can be used to gain insights into the importance of input data, including multiple variables. \cite{Tipirneni2022STraTS}, inspired by \cite{choi2016} and \cite{zhang2020}, proposed an interpretable version of the STraTS model, called STraTS-I, which uses an almost identical structure to their STraTS model but instead uses encoded inputs directly to the output layer as opposed to STraTS that uses the contextualized inputs to the output layer. Their approach allows for the calculation of a contribution score for each input observation towards the prediction, achieved by multiplying the encoded input, attention weights, and output layer weights. This modification aims to compensate accuracy for interpretability while both models have similar computational complexity. We follow a similar simplification routine in the creation of our SST-ANN approach explained in Section \ref{stmlp}.

%--------------------------------Proposed model----------------------------------------

\section{Proposed Methods}\label{model}
In this section we first define the problem followed by the details of the general model architectures that we use to build SERT and SST-ANN to address the problem. We then introduce a modification for encoding location information in the models' input data. Finally, we describe the masked loss function that we use for training the models. 

\subsection{Problem Definition}\label{problem}
\begin{figure}[htbp]
\centerline{\includegraphics[width=0.8\textwidth]{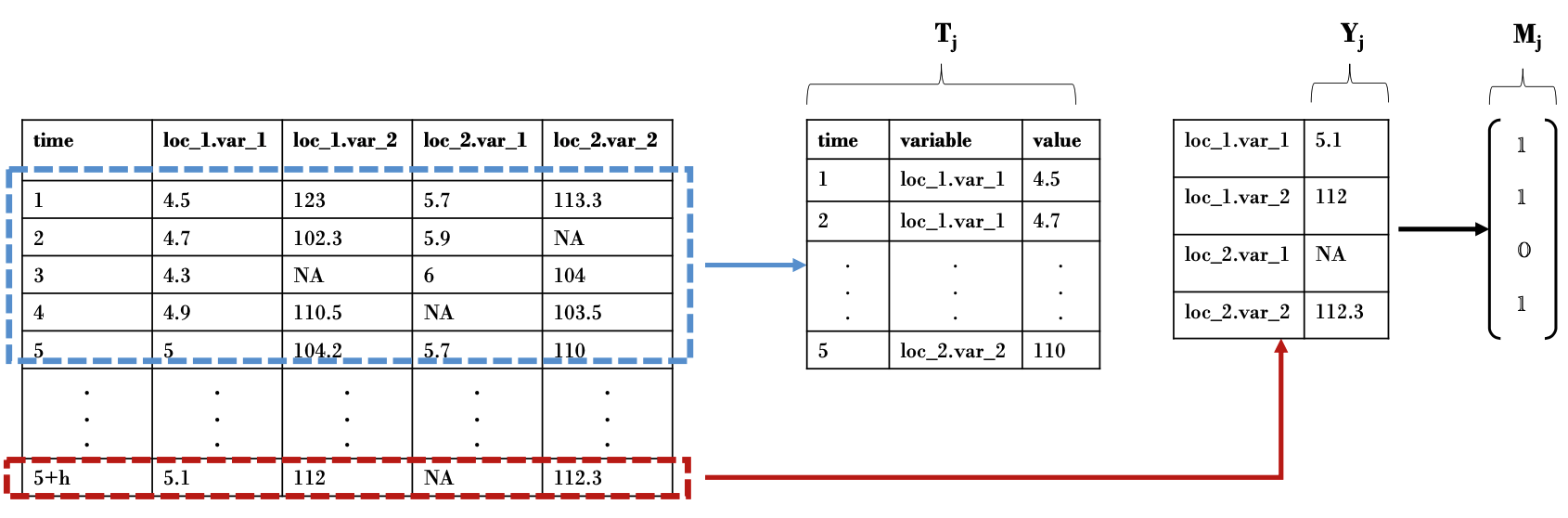}}
\caption{An example of a sample of a spatiotemporal dataset to be used for training our proposed models where $h$ is the desired forecast horizon.}
\label{data_scheme}
\end{figure}

We have a dataset $D = \{(T_{j}, Y_{j}, M_{j})\}_{j=1}^J$ where $T_{j}$ is a multivariate time series consisting of triples (time, variable, value) written as $\{(t_{i}, f_{i}, v_{i}), i = 1,\ldots,N\}$. $Y_{j}$ is the values (and associated variables) for a future time horizon for which we want the model to forecast. $M_{j}$ is a binary vector indicating whether each of element in $Y_{j}$ is observed for sample $j$ in the dataset. $M_{j}$ is used in the loss function (see Section \ref{loss} for details) for masking the unobserved values in the forecast window. A schematic of a sample from the dataset is shown in Figure \ref{data_scheme}. The goal is to learn a model $G$ that maps $T_{j}$ to $Y_{j}$, i.e., $G(T_{j}) = Y_{j}$, without imputing the missing values in $T_{j}$ or aligning the time series.

\subsection{SERT}\label{sert}
\begin{figure}[htbp]
\centerline{\includegraphics[width=1\textwidth]{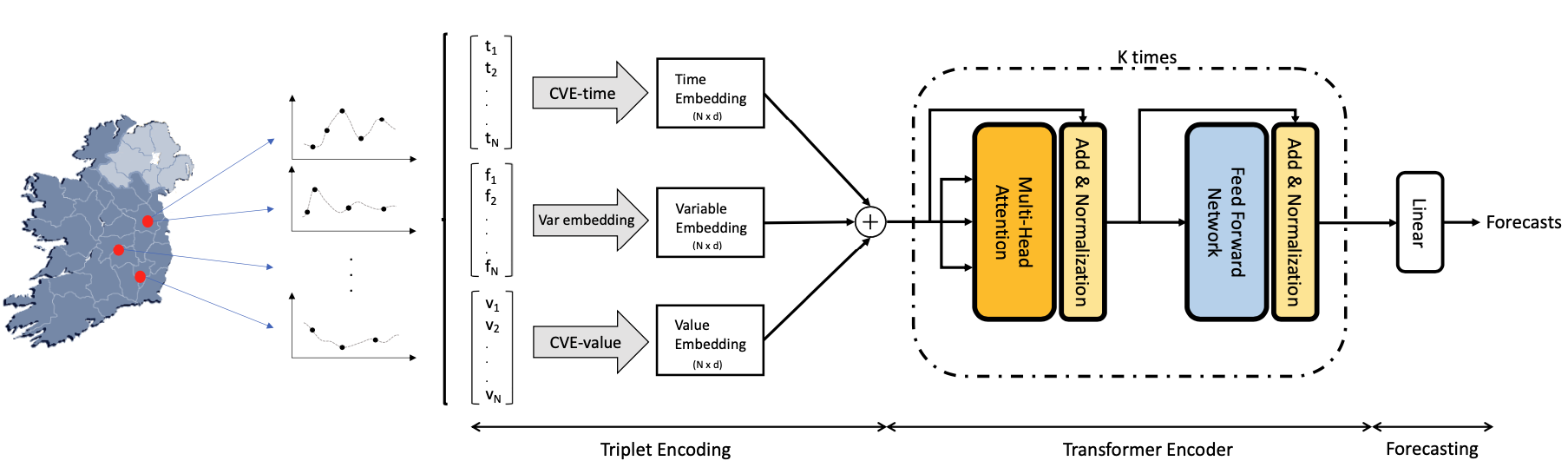}}
\caption{Schematic diagram of the SERT model.}
\label{fig:sstbert}
\end{figure}

We first describe the data encoding scheme and then the model architecture including an encoder network and a linear layer. The schematic diagram of the model is shown in Figure \ref{fig:sstbert}. We will slightly modify this structure to show an alternative approach for encoding the location information in Section \ref{location}. 

\begin{description}

\item[Data Encoding Scheme] The input data to our model is a multivariate time series dataset $T$ that consists of $N$ time series, each of which is associated with a time series variable $f$ which is a sequence of observation values $v$'s. Accordingly, an individual data point $i$ is represented as a triplet $(t_{i}, f_{i}, v_{i})$ where $t_{i}$ is time, $f_{i}$ is the variable name and $v_{i}$ is the measured value of the data point. To use the triplet in the model, we encode each component into an embedding and then add the embeddings together. Let $e_{i}^{f} \in R^{d}$ be the embedding of the variable name $f_{i}$ which can be encoded similar to words using a lookup table, $e_{i}^{t} \in R^{d}$ be the embedding of the time index $t$ which can be encoded using a continuous value embedding (CVE) scheme which is a one-to-many feed forward neural network \citep{Tipirneni2022STraTS} and $e_{i}^{v} \in R^{d}$ be the embedding of the value $v_{i}$ which can also be encoded using the CVE. The embedding of the triplet $i$ is then defined as $e_{i} = e_{i}^{f} + e_{i}^{t} + e_{i}^{v}$. The size of the embedding vector $d$ is a hyperparameter of the model. 

\item[Encoder Network] Similar to the well-known BERT model \citep{Devlin2019bert}, the main component of our model is the encoder part of the transformer model introduced by \cite{vaswani2017}. Since transformers have become very common, we omit the details of the architecture and refer the reader to \cite{vaswani2017} for the full description. Intuitively, we can think of the encoder network as layers that take the triplet embeddings of the input data and transform them into contextualized embeddings that capture the long-range dependencies within a time series as well as cross dependencies between different time series.

\item[Linear Layer] After obtaining the contextualised embeddings of the input data using the encoder network, we then flatten the embeddings and apply a linear layer to them to generate the predictions. The linear layer is a feed-forward network with a single hidden layer and a ReLU activation function. 

\end{description}

\subsection{SST-ANN}\label{stmlp}
The SST-ANN model is a simplified version of the SERT model that consists of only the triplet encoding and a linear layer to the output with no transformer structure in between. SST-ANN first encodes the input data using the triplet encoding scheme and then uses the embeddings as the input to single layer feed forward network to generate the predictions. Since there is no transformer structure in between, the SST-ANN model is much faster than the SERT model and, using the embeddings and the weights of the linear layer, we can compute a contribution score for each observation to the final prediction. This is useful for interpretability and variable importance analysis. More formally the output of the model can be expressed as follows:

\begin{equation}
    \hat{y}_{k} = \sum_{i=1}^{N} c_{i} + b,\mbox{ with } c_{i} = W_{ik}^{T} \cdot e_{i},
\end{equation}
where $\hat{y}_{k}$ is the prediction of variable $k$, $c_{i}$ is the contribution of the triplet $(t_{i}, f_{i}, v_{i})$ to the prediction, $N$ is the number of observations in the input sample, $b$ is the bias term, $e_{i}$ are the embeddings of the triplet and $W_{ik}$ is the vector of output weights associated with the embedding $e_{i}$ and the target variable $k$.

Using the contribution scores, we can define a variable importance index. We first calculate the average contribution value of all observations belonging to the same variable as the average contribution of that variable. This calculation can be performed for a single sample to gain insights into the importance of the predictor variable for a specific target prediction, or for multiple samples used in multiple predictions to obtain an overall understanding of the predictor variable's importance in general. Next, we compute the importance of each variable by normalizing the absolute value of the average contribution values for the variables. More formally, we can express this as:

\begin{equation}
    I_{k} = \frac{|\bar{c}_{k}|}{\sum_{k=1}^{K} |\bar{c}_{k}|} \times 100
\end{equation} 
where $I_{k}$ is the importance (in percentage) and $\bar{c}_{k}$ is the average contribution value of the variable $k$. 

\subsection{Location Encoding}\label{location}

\begin{figure}[htbp]
\centerline{\includegraphics[width=0.6\textwidth]{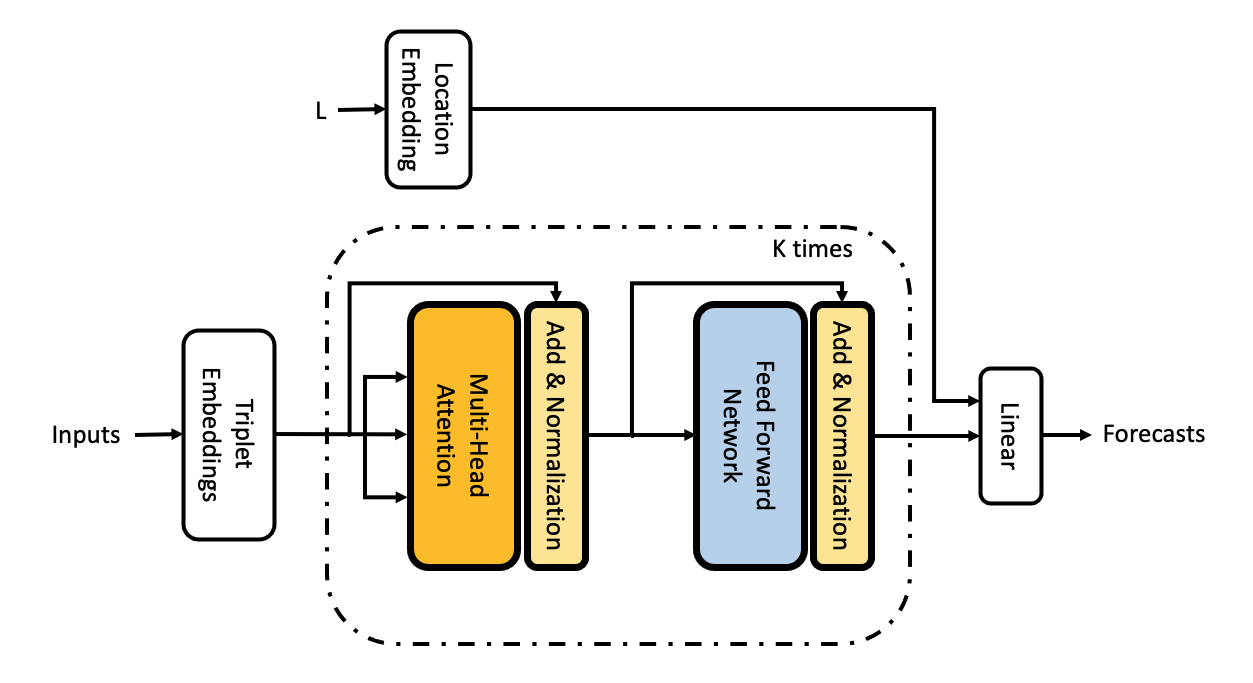}}
\caption{Schematic diagram of the SERT model with a separated location embedding layer.}
\label{new_sert}
\end{figure}

We consider two different approaches to encode the location information in the input data. The first approach is to encode the location information together with the variable name $f_{i}$ in the triplet encoding scheme. For example, our naming scheme for the variables can be $f_{i} = \{location\}_{i} \cdot \{variable\}_{i}$ where $\{location\}_{i}$ is the location of the time series $i$ and $\{variable\}_{i}$ is the variable name of the time series $i$ (e.g. Tolka.Turbidity). This approach uses the exact same architecture explained in Section \ref{sert}. The second approach is to encode the location information separately from the variable name. This way the input time series are all assumed to arise from the same location and the location embedding is concatenated to the contextualized embeddings before the linear layer is used for prediction. In this approach, we need to structure the dataset such that all the time series from the same location are grouped together. Formally, we can define the grouped dataset as $D^{\prime} = \{(T_{j}^{L}, Y_{j}^{L}, M_{j}^{L})\}_{j=1}^{J^{\prime}}$ where $L \in S$ is the location and $S$ is the set of all locations. This approach needs a minimal modification to the previously described architecture and its schematic diagram is shown in Figure \ref{new_sert}. 

The first approach is similar to how the Spacetimeformer model \citep{grigsby2019spatiotemporal} encodes the location information while the second approach is similar to how the STraTS model \citep{Tipirneni2022STraTS} encodes the non-temporal information (patient demographics in their work).

\subsection{Masked Loss Function}\label{loss}
We use a masked Mean Squared Error (MSE) loss function to train our models. Masking in the loss function is used to handle the missing values in the output data. The masked MSE loss function is defined as follows:

\begin{equation}
    \mathcal{L} =\frac{1}{\left|J\right|} \sum_{j=1}^{J} \sum_{k=1}^{K} m_k^j\left(\tilde{\mathbf{y}}_k^j-\mathbf{y}_k^j\right)^2
\end{equation}

where $\mathbf{y}_k^j$ is the ground truth, $\tilde{\mathbf{y}}_k^j$ is the predicted value, and $m_k^j$ is the mask value of the target variable $k$ in sample $j$.

%--------------------------------Experiments----------------------------------------

\section{Experiments}\label{experiments}
In this section, we will describe the experiments we conducted to evaluate the performance of our proposed models. We evaluated our models using both a simulated dataset and a real-world dataset. The primary objective of the simulation experiment is to investigate how the models perform under different sparsity levels. The real-world experiment aimed to serve as a proof of concept for the models' ability to conduct spatiotemporal multi-step ahead forecasting in real-world scenarios.

To assess the effectiveness of our models, we compared them with a baseline Naive forecaster model, which simply uses the present hour's observation as the next hour forecast. We also compare our models against the LSTM model and the STraTS model. Since the Naive forecaster and LSTM model cannot handle missing values, we first imputed these values using a forward filling method \cite[p.~16]{van2018flexible}. The source code for our experiments is available at \href{https://github.com/Aminsn/SERT2023}{https://github.com/Aminsn/SERT2023}.

\subsection{Sparsity Analysis}

We first simulated a dataset that consists of 16 time series each with 40,000 observations. We denote $Y_{t} \in R^{16}$ as the vector of observations at time $t$ generated from the following process:

\begin{equation}
     Y_{t} = 2 + 0.4 Y_{t-1} + X_{t} + s_{t},\; s_{t} \sim MVN(0, \Sigma)
\end{equation}

\noindent where $s_{t}$ is spatial random effect with mean zero and variance-covariance matrix $\Sigma$ that is generated with: 

\begin{equation}
    \Sigma = U \cdot U^{T}, \ U \in R^{16 \times 16},\; U_{i,j} \sim Uniform(-1, 1),
\end{equation}

\noindent and $X_{t} \in R^{16}$ is the vector of temporal effects generated as:

\begin{eqnarray}
    X_{t}^T &=& \left[ 10\sin(p_1t), \cos(p_2t), p_3t, -p_3t+10\sin(p_1t), 
    5\sin(p_2t), 
    12\cos(p_2t), 
    7\sin(p_2t) , 
    8\cos(p_2t), 
    %\end{bmatrix} \oplus \begin{bmatrix}
    \right. \nonumber \\
    & & \left. 2\sin(p_2t),  3\cos(p_2t), 
    12\sin(p_2t), 
    18\cos(p_2t), 
    4\sin(p_2t), 
    15\cos(p_2t), 
    11\sin(p_2t), 
    10\cos(p_2t) \right]
\end{eqnarray}

\noindent where $p_1 = 0.005$, $p_2 = 0.0005$ and $p_3 = 0.002$.

\begin{figure}[h]
    \centering
    \includegraphics[width=0.7\textwidth]{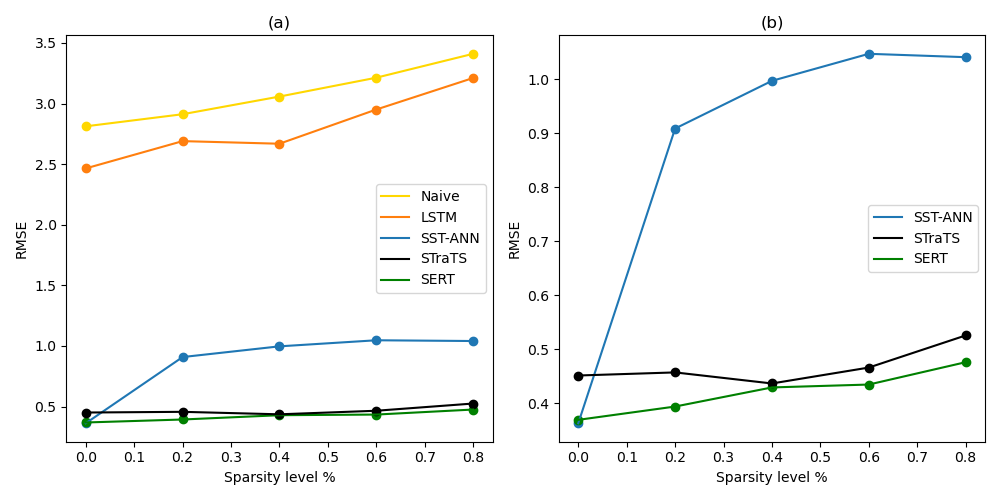}
    \caption{a) Performance of all the models used on the simulated dataset.
b) Comparison of SST-ANN, STraTS, and SERT zoomed in. }
    \label{fig:simulated}
\end{figure}

We use the first 37,000 time steps to train the models and the remaining 3,000 time steps to evaluate the performance of them. We used the model structure shown in Figure \ref{fig:sstbert} to train our proposed models. We trained all models for 1 step ahead forecasting using the previous 10 hours of observations. We consider five different sparsity levels to fit the models. Accordingly, we remove $n\%$ of the observations randomly for $n = \{0\%, 20\%, 40\%. 60\%, 80\%\}$. We use the root mean squared error (RMSE) as the evaluation metric. The results are presented in Figure \ref{fig:simulated}.

\subsection{Real Dataset; Environmental Monitoring in Dublin bay, Ireland}

\begin{figure}[htbp]
\centerline{\includegraphics[scale=.5]{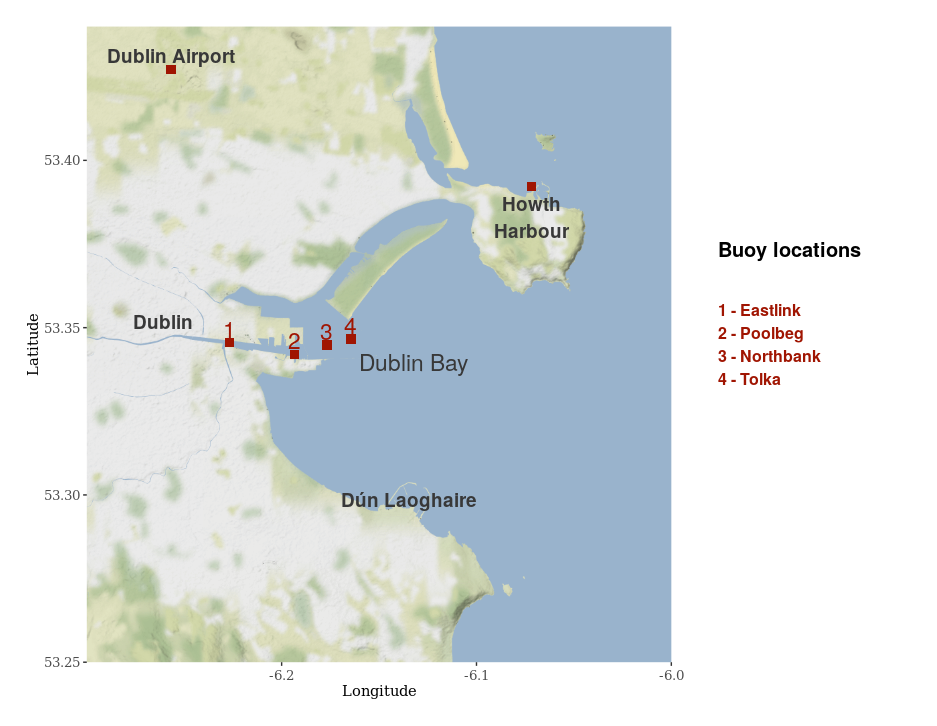}}
\caption{Buoys measuring various environmental variables in Dublin Bay. The 4 main environmental variables (Turbidity, Salinity, Dissolved Oxygen, Temperature) are available at the number buoys. The weather measurements (Rainfall, Wind Speed) come from Dublin Airport. The water level variable comes from Howth Harbour.}
\label{realworld-data}
\end{figure}

The dataset includes hourly measurements from 2017-01-01 to 2021-12-31. An example of the data is shown in Figure \ref{fig:rawdata}. The locations of the data are shown in Figure \ref{realworld-data}. We use the data from the first four years to train the models and the last year to evaluate their performance. We train the models using the previous 10 hours of observations as the input and forecasting seven hours ahead (forecasting only 1 or 2 hours ahead is a relatively trivial task, and forecasting >12 hours ahead reduces the performance of models that cannot account for non-seasonality). We use the same evaluation metric as in the simulated dataset. We tried both location encoding approaches (explained in Section \ref{location}) with our proposed models on the real-world dataset and found that the second approach, depicted in Figure \ref{new_sert}, performed better and here we only report the results of the superior approach. The results are presented in Table \ref{tab:realworld}.

\begin{table}[h!]
\centering
\begin{tabular}{|c|c|c|c|c|c|c|c|}
\hline
\textbf{Model} & \textbf{Dissolved Oxygen} & \textbf{Precipitation} & \textbf{Salinity} & \textbf{Temperature} & \textbf{Turbidity} & \textbf{Water Level} & \textbf{Wind Speed} \\
\hline
Naive & 0.64 & 1.19 & 1.22 & 0.21 & 0.93 & 1.87 & 0.8\\
\hline
LSTM & 0.65 & 0.89 & 0.8 & 0.41 & 0.81 & 0.5 & 0.82\\
\hline
STraTS & 0.53 & \textbf{0.88} & 0.7 & \textbf{0.18} & \textbf{0.72} & \textbf{0.38} & 0.76 \\
\hline
SERT (ours) & \textbf{0.49} & \textbf{0.88} & \textbf{0.67} & \textbf{0.18} & \textbf{0.72} & 0.4 & \textbf{0.73} \\
\hline
SST-ANN (ours) & 0.51 & \textbf{0.88} & 0.73 & 0.45 & 0.88 & 0.63 & 0.79 \\
\hline
\end{tabular}
\caption{RMSE of the models for 7 hour ahead forecasting of the 7 environmental variables in Dublin bay.}
\label{tab:realworld}
\end{table}

We utilized the same computational resources (a single NVIDIA P100 GPU) to train the models. The specifications and speed performance details of the models are reported in the Table \ref{speedtbl}. 

\begin{table}[h!]
\centering
\begin{tabular}{|c|c|c|}
\hline
\textbf{Model} & \textbf{Specifications} & \textbf{Sec/epoch} \\
\hline
Naive & --- & --- \\
\hline
LSTM & units = 60 & 2\\
\hline
STraTS & d = 60, \# of heads = 6, k = 6 & 230\\
\hline
SERT & d = 60, \# of heads = 6, k = 6 & 40\\
\hline
SST-ANN & d = 60 & 9 \\
\hline
\end{tabular}
\caption{Computational specifications of the fitted models.}
\label{speedtbl}
\end{table}

\subsection{Interpretability of the SST-ANN model}\label{interpretability}
\begin{figure}[h]
    \centering
    \includegraphics[width=0.5\textwidth]{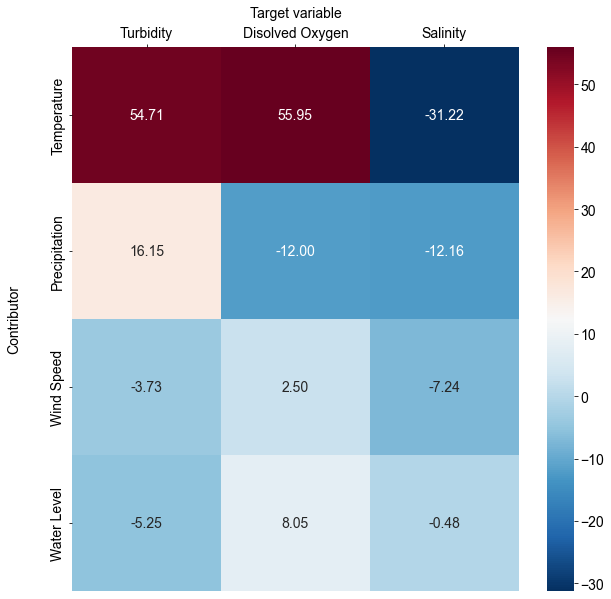}
    \caption{Variable importance of the selected variables in predicting the target variables.}
    \label{fig:realworld-importance}

\end{figure}

The results of the variable importance for the real-world dataset experiment are presented in Figure \ref{fig:realworld-importance}. The sign of the average contribution score $\bar{c}_{j}$ (explained in Section \ref{stmlp}) is multiplied by the importance index $I_{j}$ to give an insight into the direction of the average contribution of the predictor variable to the prediction of the target variable. In this example, we only consider the contributions of water level, temperature, wind speed and precipitation to predictions of turbidity, disolved oxygen and salinity, since we know that the former variables could affect the latter variables but not vice versa. According to the results, temperature followed by precipitation are the most important variables in predicting the target variables.

%--------------------------------Conculusion and Discussion----------------------------------------

\section{Conclusion}\label{conclusion}
In this paper, we proposed two novel models for spatio-temporal forecasting called SERT and SST-ANN. SERT is a transformer based model while SST-ANN is a simple ANN model combined with triplet encoding of the input data. Furthermore, we showed that STraTS, a model originally developed for sparse and irregularly sampled clinical time series classification, can be used for spatio-temporal forecasting, especially when missing values are present in the data. The proposed approaches do not require aggregation or missing value imputation techniques, and avoid the problems introduced by such methods. We evaluated the performance of the proposed models on a simulated dataset with varying levels of sparsity and showed that in general increasing sparsity has a negative effect in the performance of all the models, but SERT followed by STraTS and SST-ANN are more robust to the increase in sparsity. We also evaluated the performance of the proposed models on a real-world dataset of environmental variables in Dublin Bay, Ireland. The results indicate that SERT outperformed the other models in 7-hour ahead forecasting for 6 out of the 7 variables, with 3 of them being on par with STraTS while demonstrating significantly faster performance. We then showed how SST-ANN can be used to interpret the predictions of the model by calculating and using the contribution score of the input data to develop an importance index using the average contribution scores.

We introduced two different methods to encode the location information in our proposed models, including encoding the time series variable name with the location name simultaneously and encoding the location name separately. However, neither of these methods takes into account the distance between the locations, which is a limitation of our work. We believe future research should focus on incorporating this information into the models, as it has the potential to improve forecasting performance and be utilized for spatiotemporal interpolation tasks.

\section*{Acknowledgement}

We would like to express our sincere gratitude to Dublin Port Company for providing us with the real dataset. This work was supported by an SFI Investigator award (16/IA/4520).

\bibliography{wileyNJD-APA.bib}

\end{document}